\pdfoutput=1

\documentclass[11pt]{article}

\usepackage[]{acl}

\usepackage{times}
\usepackage{latexsym}

\usepackage[T1]{fontenc}

\usepackage[utf8]{inputenc}

\usepackage{microtype}

\usepackage{inconsolata}

\usepackage{algorithmic}
\usepackage{newfloat}
\usepackage{listings}
\DeclareCaptionStyle{ruled}{labelfont=normalfont,labelsep=colon,strut=off} 
\usepackage{inconsolata}
\usepackage{calligra}
\usepackage{helvet}
\usepackage{newtxtt}
\usepackage{xcolor}
\usepackage{amssymb}
\usepackage{amsxtra}
\usepackage{multirow}
\usepackage{diagbox}
\usepackage{longtable}
\usepackage{array}
\usepackage{stfloats}
\usepackage{graphicx}
\usepackage{multicol}
\usepackage{amsmath, bm}
\usepackage{booktabs} 
\usepackage{arydshln}
\usepackage{xspace}
\usepackage{enumitem}
\usepackage{threeparttable}
\usepackage{dsfont}
\usepackage{bbm}
\usepackage{makecell}
\usepackage[linesnumbered,ruled,vlined]{algorithm2e}
\usepackage{color} 
\usepackage{subfigure}
\newcommand{\hide}[1]{}

\newcommand{\vpara}[1]{\vspace{0.05in}\noindent \textbf{#1 }}

\newcommand{\beq}[1]{\vspace{-0.03in}\begin{equation}#1\end{equation}\vspace{-0.03in}}

\newcommand{\red}[1]{\textcolor{red}{#1}}
\title{SGSH: Stimulate Large Language Models with Skeleton Heuristics for Knowledge Base Question Generation}

\author{Shasha Guo\textsuperscript{1, 2}, 
 Lizi Liao\textsuperscript{3}, Jing Zhang\textsuperscript{1, 2}\thanks{       $\text{ }$ \scriptsize Corresponding author. This work was done during an internship at SMU.\normalsize}, Yanling Wang\textsuperscript{4}, \\
\textbf{Cuiping Li\textsuperscript{1, 2}}, \textbf{Hong Chen\textsuperscript{1, 2}} \\
  \textsuperscript{1}School of Information, Renmin University of China, Beijing, China\\
  \textsuperscript{2}Key Laboratory of Data Engineering and Knowledge Engineering of Ministry of Education\\
  \textsuperscript{3}Singapore Management University \textsuperscript{4}Zhongguancun Laboratory\\
  \{guoshashaxing, zhang-jing, licuiping, chong\}@ruc.edu.cn\\
  lzliao@smu.edu.sg, wangyl@zgclab.edu.cn
  }

\begin{document}

\maketitle

\begin{abstract}

Knowledge base question generation (KBQG) aims to generate natural language questions from a set of triplet facts extracted from KB. Existing methods have significantly boosted the performance of KBQG via pre-trained language models (PLMs) thanks to the richly endowed semantic knowledge.
With the advance of pre-training techniques, large language models (LLMs) (e.g., GPT-3.5) undoubtedly possess much more semantic knowledge.
Therefore, how to effectively organize and exploit the abundant knowledge for KBQG becomes the focus of our study.
In this work, we propose \textbf{SGSH} --- a simple and effective framework to \underline{\textbf{S}}timulate \underline{\textbf{G}}PT-3.5 with \underline{\textbf{S}}keleton \underline{\textbf{H}}euristics to enhance KBQG. 
The framework incorporates ``\textit{skeleton heuristics}'', which provides more fine-grained guidance associated with each input to stimulate LLMs to generate optimal questions, encompassing essential elements like the question phrase and the auxiliary verb.
More specifically, we devise an automatic data construction strategy leveraging ChatGPT to construct a skeleton training dataset, based on which we employ a soft prompting approach to train a BART model dedicated to generating the skeleton associated with each input.
Subsequently, skeleton heuristics are encoded into the prompt to incentivize GPT-3.5 to generate desired questions. 
Extensive experiments demonstrate that SGSH derives the new state-of-the-art performance on the KBQG tasks.
The code is now available on Github\footnote{\href{https://github.com/RUCKBReasoning/SGSH}{https://github.com/RUCKBReasoning/SGSH}}.

\end{abstract}
\section{Introduction}
\label{sec:intro}
Knowledge Base Question Generation (KBQG) has attracted a lot of attention owing to its wide range of applications in academia and industry~\cite{guo2024survey}. 
On the one hand, KBQG can augment training data for question answering (QA) to improve the performance of QA models~\cite{chen2020toward, DSM}.
On the other hand, KBQG empowers machines to actively ask questions in conversations with humans~\cite{ConvMR, MWPG}.

\begin{figure}[!t]
\centering 
\resizebox{0.45\width}{0.425\height}{\includegraphics{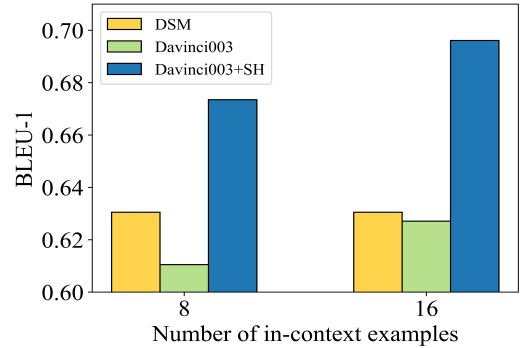}}
\caption{Performance comparison between three advanced methods for KBQG under different numbers of in-context examples on the  WebQuestions dataset.
The methods include the state-of-the-art PLM-based method DSM (yellow), text-davinci-003 (green), and text-davinci-003 with skeleton heuristics (blue).}
\label{fig:pilot_study} 
\end{figure}
 
The quality of KBQG has been significantly improved,
largely attributable to the success of pre-trained language models (PLMs) like BART~\cite{BART} and T5~\cite{T5}. 
A noteworthy example is DSM~\cite{DSM}, which introduces a meta-learner based on the BART for KBQG, effectively capturing diverse semantic information within a KB. Moreover, AutoQGS~\cite{AutoQGS} designs an auto-prompt approach upon the BART, achieving the low-resource KBQG.
Current PLMs, pre-trained on comprehensive corpora, come equipped with rich semantic knowledge, facilitating significant performance improvements in the downstream KBQG task upon fine-tuning.

Recently, large language models (LLMs) such as InstructGPT~\cite{InstructGPT} and ChatGPT\footnote{https://openai.com/blog/chatgpt}, have exhibited impressive capabilities in a variety of tasks~\cite{codeGeneration, Text-to-SQL}.
However, the vast amount of generalized knowledge poses a challenge in extracting pertinent information for the KBQG task, making LLMs fall short of the expected performance on KBQG.
As demonstrated in Figure~\ref{fig:pilot_study}, the PLM-based cutting-edge approach DSM outperforms the direct application of the LLM, Davinci003.
In view of this, our focus is on how to trigger LLMs to effectively utilize their knowledge to improve the quality of KBQG, which is an under-explored problem in the community of natural language processing.

Inspired by the way humans learn a language, which typically involves acquiring grammatical knowledge before progressing to reading and writing, we summarize the grammatical elements to guide the desired question generation, instead of directly applying an LLM.
In this work, the grammatical elements include the question word phrase and the auxiliary verb, which we call ``\textit{skeleton}''.
Through a pilot study, we observe that prompts coupled with skeleton heuristics can boost the performance of Davinci003 on the KBQG task (Cf. the comparison between Davinci003 and Davinci003+SH in Figure~\ref{fig:pilot_study}).
In effect, the skeleton heuristics can be viewed as the fine-grained guidance to excavate task-specific knowledge from the LLMs, thereby stimulating the LLMs to generate more accurate questions.

Motivated by the above insights, we propose  \textbf{SGSH} --- a
simple and effective framework to \underline{\textbf{S}}timulate \underline{\textbf{G}}PT-3.5\footnote{We use \texttt{text-davinci-003} and \texttt{gpt-3.5-turbo}.} with \underline{\textbf{S}}keleton \underline{\textbf{H}}euristics for KBQG, which contains two modules, i.e., a skeleton generator and a black-box LLM (e.g., GPT-3.5).
Figure~\ref{fig:framework} illustrates the overview of SGSH.
Specifically, a skeleton generator implemented by a small PLM (e.g., BART) generates the skeleton for each input, where the skeleton is a series of discrete tokens that act as a particular signal to guide the LLM toward the ground-truth question.
To train the skeleton generator, we propose an automatic strategy to construct a high-quality training dataset, which leverages a rule-based method to initially extract skeletons and then utilizes the power of ChatGPT to refine these skeletons.
Based on the training set, we learn the skeleton generator with a soft prompting strategy to generate the skeleton for each input.
Subsequently, the black-box LLM utilizes the skeleton heuristics via \textit{skeleton injection} and \textit{skeleton-aware in-context learning}.
Concretely, given a test input consisting of triples along with the corresponding answer, the skeleton injection step integrates the generated skeleton into the test input.
Afterward, the skeleton-aware in-context learning step incorporates in-context examples with skeletons to effectively enhance the in-context learning capability for the test input, where each example shares a similar target question with the test input.

\vpara{Key Contributions.} 1) The development of an automatic data-building approach with a soft prompting strategy for effective skeleton heuristic generation. 2) The creation of an enhanced prompting mechanism, with skeleton injection and skeleton-aware in-context learning, steers GPT-3.5 towards generating more precise questions. 3) Demonstrated superiority of our approach over existing methods in both automatic and human evaluations, also proving beneficial for data augmentation in question answering tasks.

\section{Pilot Study}
\label{sec:pilot}

To evaluate the effectiveness of the skeleton heuristics in enhancing the performance of
KBQG, we undertake a preliminary investigation to analyze.

\vpara{KBQG.} Given a set of triples extracted from a KB and a particular answer, the objective of KBQG is to generate a question associated with the answer. $\mathcal{D} = \{(G_i, a_i, q_i)\}_{i=1}^N$ denotes the dataset for training a KBQG model, where $G_i$ represents a subgraph comprising a set of triples, $a_i$ signifies a given answer, and $q_i$ denotes the target question. 
This research explores the use of black-box LLMs like GPT-3.5 for KBQG, which can only be accessed through APIs.

\vpara{Modeling.}
We perform a pilot study on the commonly used KBQG benchmark, WQ~\cite{WQ, CWQ}.
To reduce the cost of API usage, we randomly sample 50 test examples $\{(G_j, a_j)\}_{j=1}^{50}$ from the WQ  test set for evaluation.
The existing state-of-the-art PLM-based KBQG method DSM~\cite{DSM} is our baseline.
Another baseline is directly using Davinci003 for KBQG, which takes the test example $(G_j, a_j)$ as input and predicts the corresponding question.
To investigate the potential benefits of skeleton heuristics, we use a skeleton generator to derive skeletons for the sampled 50 test examples.
To train the skeleton generator, we first construct a skeleton training dataset based on $\mathcal{D} = \{(G_i, a_i, q_i)\}_{i=1}^N$ by a rule-based method, which extracts the skeleton elements (i.e., the question word phrase and the auxiliary verb) from $q_i$ by searching a pre-defined vocabulary of skeleton elements.
Based on the skeleton training dataset, we proceed to train a skeleton generator that produces the specific skeleton heuristics for each test example $(G_j, a_j)$.
Subsequently, these elicited skeleton heuristics are seamlessly incorporated into the test input to stimulate Davinci003 to generate the desired question. The skeleton heuristics-based approach is denoted as Davinci003+SH.

\vpara{Observation --- skeleton heuristics can unlock the potential of LLMs for the KBQG task.}
Figure~\ref{fig:pilot_study} illustrates that directly applying the LLM (i.e., Davinci003) falls short in performance compared to the PLM-based method (i.e., DSM) in terms of BLEU-1 metric. 
However, Davinci003+SH, which considers the skeleton heuristics, outperforms both DSM and Davinci003.
This implies that directly employing LLMs might not fully exploit useful knowledge to generate the intended questions. 
The incorporation of skeleton heuristics can enhance the performance of LLMs, serving as an accurate guiding signal that aids LLMs in aligning their output with the gold question.
Inspired by these findings, we propose our novel approach SGSH.

\section{Methodology}
\label{sec:method}

\begin{figure*}[!t]
\centering 
\includegraphics[width=0.95\textwidth]{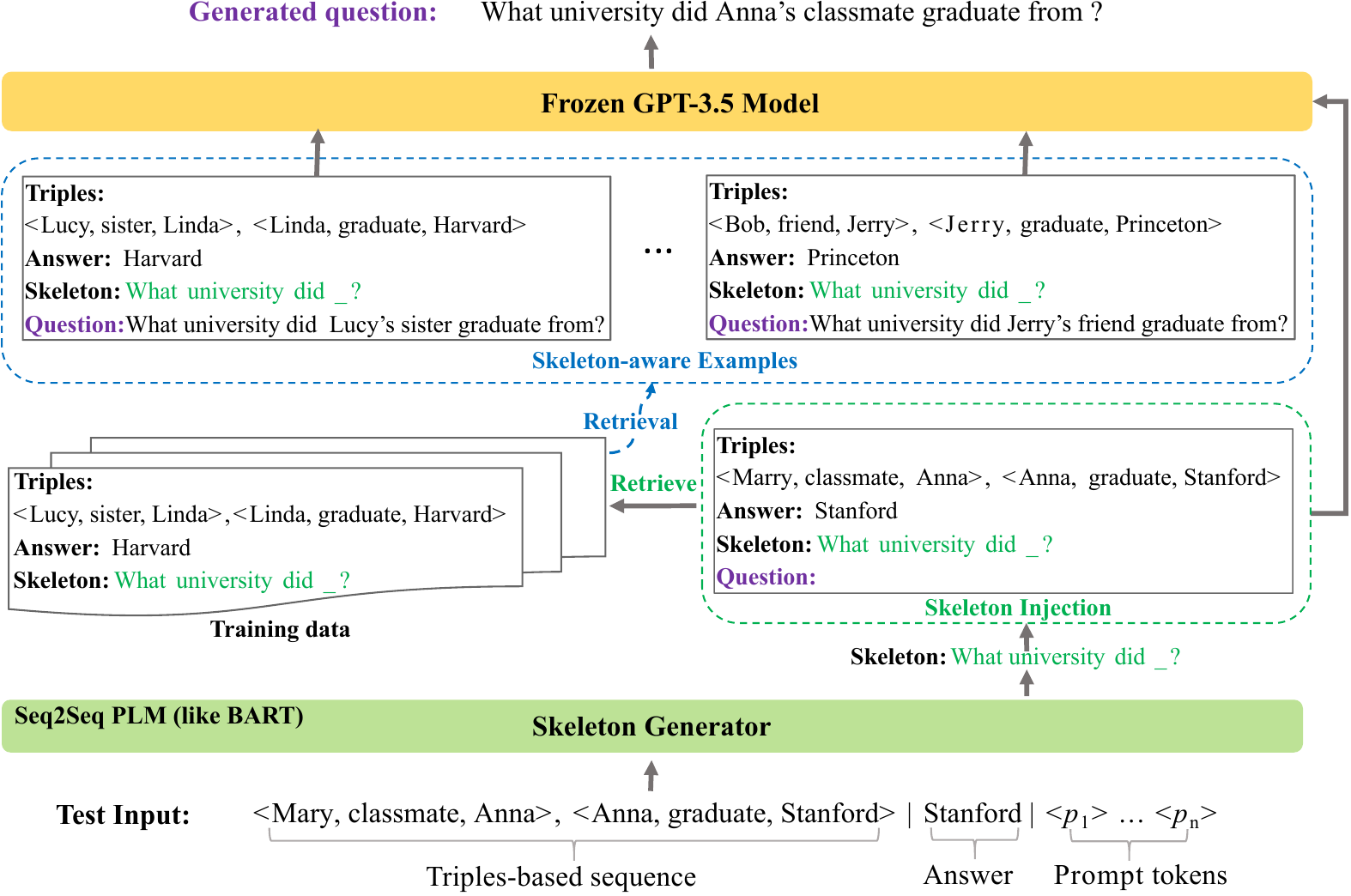}
\caption{Overview of our SGSH framework, which consists of a PLM-based skeleton generator and a frozen GPT-3.5 model. 
The skeleton generator, optimized by the learnable prompting strategy, generates the skeleton for each test input.
Subsequently, GPT-3.5 leverages skeleton heuristics through skeleton injection and skeleton-aware in-context learning to generate the desired question.
}
\label{fig:framework} 
\end{figure*}

\subsection{Model Overview}
Our proposed SGSH framework comprises two pivotal modules, a PLM-based skeleton generator (e.g., BART) and a frozen GPT-3.5 model (e.g., Davinci003). 
Figure~\ref{fig:framework} illustrates the overall framework.
The skeleton generator produces skeletons of test inputs as a precise signal to steer GPT-3.5 at a fine-grained level. 
More specifically, we first construct a skeleton training dataset $\mathcal{D_S} =  \{(G_i, a_i, s_i)\}_{i=1}^N$ based on $\mathcal{D} = \{(G_i, a_i, q_i)\}_{i=1}^N$ leveraging an automatic data construction strategy and then use a small tunable PLM to learn from the obtained training dataset $\mathcal{D_S}$.
Subsequently, GPT-3.5 employs skeleton heuristics through skeleton injection and skeleton-aware in-context learning. 
These strategies steer GPT-3.5 to skillfully leverage its internal knowledge, thus enhancing its effectiveness in advancing the KBQG task.

\subsection{Skeleton Generator}
We explain how to (1) perform \textbf{an automatic training data construction} strategy to acquire supervised data and (2) \textbf{fine-tune} the skeleton generator with learnable prompting to produce skeletons.

\begin{figure}[!t]
\centering 
\includegraphics[width=0.45\textwidth]{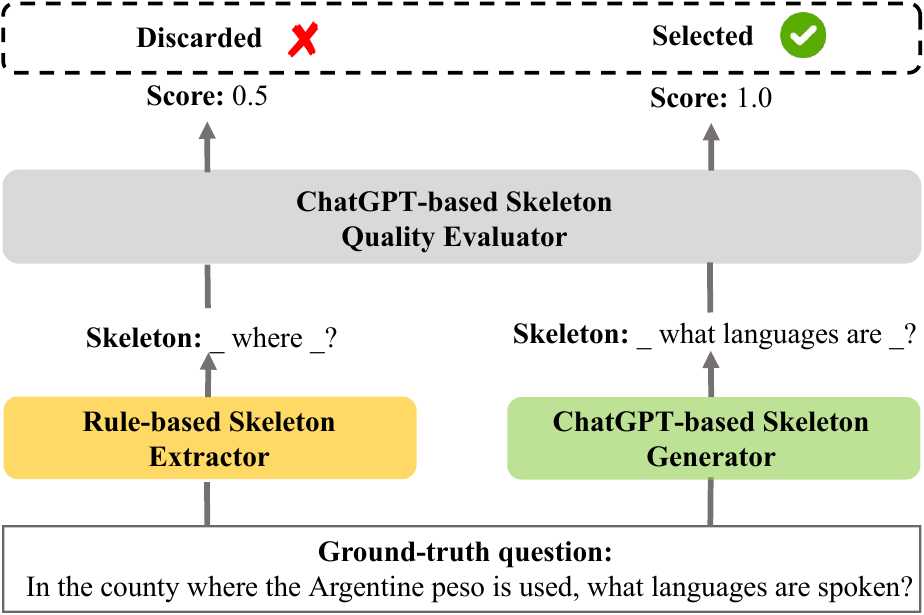}
\caption{Illustration of the automatic training data construction strategy. We use ChatGPT as an automatic scorer to rate
each skeleton generated by the rule-based and ChatGPT-based methods on a scale of 0 to 1.
}
\label{fig:data_building} 
\end{figure}

\vpara{Automatic Training Data Construction.} In order to effectively train a skeleton generator through supervised fine-tuning, we need to collect labeled data.
To avoid costly and time-consuming human annotation, we devise an automatic approach to construct the required data.
Drawing inspiration from human cognitive processes, the essential elements of a question are the question word phrase and the auxiliary verb, which are defined as the ``\textit{skeleton}''.
Specifically, we first derive skeletons from $\mathcal{D} = \{(G_i, a_i, q_i)\}_{i=1}^N$ using a rule-based method, which extracts the skeleton elements from $q_i$ by searching a pre-defined vocabulary of skeleton elements.
Considering the limitation of the rule-based method, such as difficulty in solving complex questions with nested clauses (Cf. Figure~\ref{fig:data_building}), and the powerful capabilities of ChatGPT, we utilize ChatGPT to generate skeletons with a well-designed prompt.
Subsequently, we employ ChatGPT as an automatic grader to score the skeletons obtained through the aforementioned methods. We then select the skeleton with the comparatively higher score as the definitive one.
By doing this, we obtain the supervised data $\mathcal{D_S} =  \{(G_i, a_i, s_i)\}_{i=1}^N$ consisting of input-skeleton pairs, which are used to train the skeleton generator to infer skeletons for test inputs without requiring any additional manual labeling.
Figure~\ref{fig:data_building} shows the overall pipeline.

\vpara{Fine-tuning a PLM-based Skeleton Generator.} 
To train the skeleton generator, one straightforward method is to perform vanilla fine-tuning on a tunable PLM $f_{\text{PLM}}$ (i.e., BART\footnote{https://huggingface.co/facebook/bart-base}).
Motivated by the Prompt Tuning work~\cite{prompt_tuning}, we enhance the vanilla fine-tuning by utilizing a learnable prompting training strategy to effectively train $f_{\text{PLM}}$ for precise alignment with the target skeleton~\cite{ClusterPrompt}.
Specifically, we linearize $G_i$ into a triple-based sequence, with each triple separated by commas.
Then we append the representations of the prompt tokens $\mathcal{P} = \{p_1, p_2, ..., p_n\} $ to the end of the input $(G_i, a_i)$, which will be updated during the training process.
Formally, the objection function is defined as:
\beq{
\label{eq:vanilla_finetune}
    \mathcal{L}(\theta, \theta_{p}) = \max_{\theta; \theta_{p}} \sum_{i=1}^{N} \log  P_{\theta; \theta_{p}}\left(s_i| G_i, a_i, \mathcal{P}\right) ,
}

\noindent where $f_{\text{PLM}}$ contains two types of parameters, $\theta$ and $\theta_{p}$. 
The former is the backbone BART parameters and the latter is the prompt specialized parameters.

Notably, training $t$ groups\footnote{In our experiments, we set the value of t as 8.} of learnable prompts, each with different hyperparameters, and subsequently ensembling them during the inference phase can significantly boost the performance of the model.
In addition, we find an intriguing phenomenon --- few supervised data (i.e., 10\%) can achieve comparable performance to full supervised data.
We validate the performance across different numbers of learnable prompt groups and supervised data in our experiments.

\subsection{Skeleton Heuristics-Enhanced Prompting}
Inspired by the observation that skeleton heuristics can stimulate GPT-3.5 for the KBQG task, we introduce skeleton heuristics into the prompt, called \textit{skeleton heuristics-enhanced prompt}, which provides more fine-grained guidance for GPT-3.5.
We elaborate on how to utilize skeleton heuristics through two distinct approaches: \textbf{skeleton injection} and \textbf{skeleton-aware in-context learning}.

\vpara{Skeleton Injection.} The one approach represents injecting the skeleton generated by a skeleton generator $f_{\text{PLM}}$ into the test input $(G_j, a_j)$.
In specific, given a test input $(G_j, a_j)$ and the corresponding skeleton $s_j$ = $f_{\text{PLM}}\left(G_j, a_j, \mathcal{P}\right)$, we can obtain the skeleton injection $(G_j, a_j, s_j)$.

\vpara{Skeleton-aware In-Context Learning.} 
The alternative approach incorporates in-context examples with corresponding skeletons (Cf. Automatic Training Data Construction) to facilitate in-context learning for test inputs.
In this method, each in-context example shares a similar target question with the test input, thereby enhancing the quality of the question generated by the test input.
Previous studies have revealed that different in-context examples may affect the performance of LLMs~\cite{in_context, in_context_gpt}. 
Motivated by this, we devise a skeleton-aware example selection strategy called \textit{input+skeleton emb}.
Concretely, given a test input skeleton injection $(G_j, a_j, s_j)$ and a training example skeleton injection $(G_i, a_i, s_i)$, we apply a small PLM (i.e., BART) trained on $\mathcal{D}$ to obtain their corresponding embeddings $e_j$ and $e_i$\footnote{Experimentally, we utilize the last hidden state of BART encoder as the embedding.}. 
Since the embedding is used to decode the target question, it contains rich semantic information about the question for the given input-skeleton pairs. 
Therefore, if $e_j$ and $e_i$ are close in the embedding space, they probably correspond to similar target questions.
Then, we calculate the cosine similarity between the test input embedding $e_j$ and each training example embedding $e_i$ in $\mathcal{D}$ and select the Top-$k$ most similar training examples as the skeleton-aware examples, i.e.,
\beq{
\label{eq:cosine}
    SE(e_j) = \underset{i\in \{1,2,..., N\}} {TopK} \frac{e_j \cdot e_i}{||e_j||_2 ||e_i||_2} .
}

The embeddings of the training examples can be calculated and stored in advance so that skeleton-aware examples can be efficiently selected.

\begin{figure}[!t]
\centering 
\includegraphics[width=0.46\textwidth]{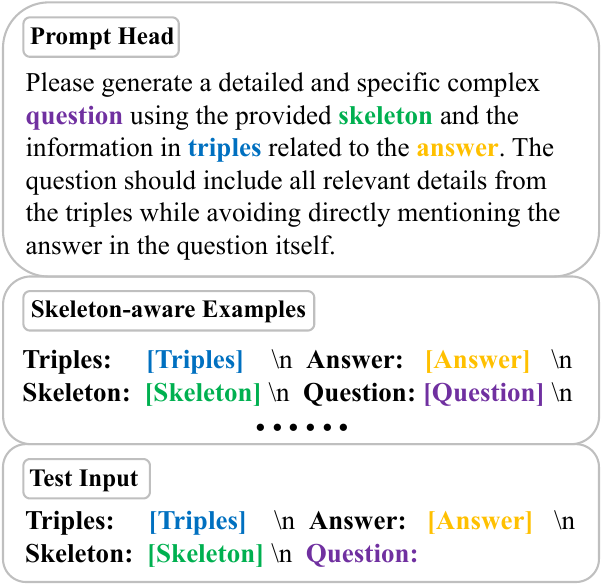}
\caption{A skeleton heuristics-enhanced prompt for Davinci003 on KBQG.}
\label{fig:prompt_format} 
\end{figure}

Figure~\ref{fig:prompt_format} illustrates the skeleton heuristics-enhanced prompt consisting of a prompt head, a set of skeleton-aware examples, and a test input with a skeleton. 
Specifically, the prompt head serves as an explanation of the KBQG task, necessitating clarity and specificity to elicit responses that meet our intended requirements.
Skeleton-aware examples are derived from the Top-$k$ skeleton-aware examples $SE(e_j)$, each containing a corresponding question similar to the target question of the test input $(G_j, a_j)$.
Notably, the number of skeleton-aware examples $k$ affects the performance of the generated question, which will be validated in the experiments (Cf. Ablation Studies~\ref{sec:ablation_study}).
The test input skeleton injection $(G_j, a_j, s_j)$ follows a similar format to skeleton-aware examples.
The only difference lies in that the question slot will be generated by the GPT-3.5 model.

\section{Experiments}
\label{sec:experiment}

\subsection{Experimental Settings}
\vpara{Datasets.}
We evaluate the proposed method on two widely used datasets WebQuestions (WQ) and PathQuestions (PQ)~\cite{PQ}.
Concretely, WQ includes 22,989 instances from WebQuestionsSP~\cite{WQ} and ComplexWebQuestions~\cite{CWQ}, which are divided into training set/dev set/test set with 18,989/2,000/2,000 instances.
PQ consists of train/dev/test set with 9,793/1,000/1,000 instances.

\vpara{Evaluation Metrics.}
For evaluation, we employ automatic evaluation metrics, human evaluation, and the downstream QA task. 
For automatic evaluation metrics, we use two classic metrics, namely BLEU-$n$ ($n$ = 1-4)~\cite{bleu} and ROUGE-L~\cite{rouge}, which calculate the proportion of identical n-grams between the generated question and the gold question.
The former can be seen as precision, while the latter focuses on recall. 
For downstream QA tasks, we report the F1 score as some questions have multiple answers. 
To measure the accuracy of the top-1 predicted answer, we use the Hits@1 metric.
For human evaluation, we invite three persons to evaluate the relevance and fluency of generated questions.

\begin{table*}[ht]
\centering
\newcolumntype{?}{!{\vrule width 1pt}}
\renewcommand\arraystretch{1.0}
\scalebox{0.71}{
\begin{tabular}{@{}c@{ }?@{ }ccccc@{ }?@{ }ccccc@{}}
\toprule
\multirow{2}{*}{\textbf{Model}} &
\multicolumn{5}{@{}c@{ }?@{ }}{\textbf{WQ}} &
\multicolumn{5}{@{}c}{\textbf{PQ}} \\
& \textbf{BLEU-1}& \textbf{BLEU-2} & \textbf{BLEU-3}& \textbf{BLEU-4}& \textbf{ROUGE-L}& \textbf{BLEU-1}& \textbf{BLEU-2} & \textbf{BLEU-3}& \textbf{BLEU-4} & \textbf{ROUGE-L}\\
\midrule
\multicolumn{11}{c}{\textbf{Non-PLMs models}}\\
\midrule
MHQG+AE & 42.35 & 29.32 & 18.43 & 9.63 & 35.72 & 45.02 &35.86 & 28.73&17.86&63.45\\
G2S+AE & 53.48 & 38.67& 27.35 & 20.54 & 55.61 &  78.21 &69.62 & 63.35&54.21 &82.32\\

G2S+AE+RL & 54.69& 39.77& 27.35 & 20.80 &55.73 & 76.05 &67.75 &  61.64&52.19 &81.94\\
\midrule
\multicolumn{11}{c}{\textbf{PLMs-based models}}\\
\midrule
T5 & 50.14 & 37.01& 28.24 & 21.88 &50.20 & 75.46 & 67.99 & 63.01& 57.79 & 75.69\\
BART & 56.39 & 41.05& 29.59 & 21.46 &56.51 & 79.59 & 70.63 &64.30& 55.73 & 84.54\\
JointGT(T5) & 55.55 & 39.71& 29.61 & 22.57 &56.23 & 77.87  &69.38 & 63.49&56.17& 81.98\\
JointGT(BART) &56.80 & 41.27& 31.23 & 24.01&57.29 & 81.67  & 72.80 &66.97&59.88& 83.61\\
DSM & 62.94 & 48.20& 37.50 &28.62 & 64.25 & 82.44 & 74.20 & 68.60& 61.03 &86.06\\
B+S & 64.44 & 52.83&44.20 & 36.70 & 67.41 & 86.57  & 79.03 &73.28& 65.63& 89.45\\
\midrule
\multicolumn{11}{c}{\textbf{GPT-3.5-based models}}\\
\midrule
ChatGPT & 56.46 &41.76& 31.92 & 24.36 &58.23& 78.45  & 70.88 & 64.88& 57.52& 84.72\\
Davinci003 & 61.68 & 47.85& 38.05 & 30.00 &61.80 & 80.53  &73.55 & 68.14& 61.67& 86.48\\

\midrule
\multicolumn{11}{c}{\textbf{Our proposed approach}}\\
\midrule
SGSH(ChatGPT) & 63.30 & 50.34& 40.89 & 32.78 &65.46 & 83.81  & 77.28 &72.04& 65.13& 87.78\\
SGSH(Davinci003) & \textbf{68.16} &\textbf{56.32}& \textbf{47.30} & \textbf{39.12} & \textbf{69.59} & \textbf{88.87}  & \textbf{83.76} &\textbf{79.52}& \textbf{74.13}& \textbf{92.47}\\
\bottomrule
\end{tabular}
}
\caption{Overall evaluation on WQ and PQ (\%).}
\label{tb:overall_evaluation}
\end{table*}

\vpara{Baselines.} 
We compare with \textbf{Non-PLMs models}, in which MHQG+AE~\cite{kumar2019difficulty} directly feeds the subgraph into Transformer~\cite{vaswani2017attention} to generate the question. 
G2S+AE and G2S+AE+RL~\cite{chen2020toward} employ a bidirectional gated GNN to encode the subgraph and use the LSTM model to decode the question, whereas the latter adds the reinforcement loss to reward the model for generating better questions.
In addition, we also compare with \textbf{PLMs-based models}, where BART~\cite{BART} and T5\footnote{https://huggingface.co/t5-base}~\cite{T5} are directly fine-tuned to solve the KBQG task. 
JointGT(BART) and JointGT(T5)~\cite{JointGT} inject the structure-aware semantic aggregation module into the vanilla PLMs to preserve the graph structure and devises three pre-training tasks to learn graph-text alignment.
DSM~\cite{DSM} focuses on the diversity of subgraphs and models the diverse subgraph via meta-learner~\cite{MAML}.
BART+Skeleton (abbreviated as B+S) represents our developed baseline, trained on the raw input and its corresponding skeleton.
Finally, we compare with \textbf{GPT-3.5-based models}, where Davinci003 and ChatGPT (i.e., \texttt{gpt-3.5-turbo}) are directly used for the task.

\subsection{Overall Evaluation}
In Table~\ref{tb:overall_evaluation}, we present the comprehensive assessment findings for WQ and PQ. Based on these findings, the following conclusions can be made: 
\textbf{(1) Directly applying GPT-3.5 to KBQG fails to achieve good performance.} 
Compared to existing state-of-the-art (SOTA) PLMs-based method (i.e., DSM), we notice that ChatGPT reduces 6.48\% BLEU-1 and 6.02\% ROUGE-L, while Davinci003 reduces 1.26\% BLEU-1 and 2.45\% ROUGE-L on WQ.
This aforementioned performance does not match the remarkable capabilities of GPT-3.5, which can be explained that employing GPT-3.5 directly with a vanilla prompt only provides coarse-grained guidance but cannot offer specific and accurate guidance direction, resulting in poor quality of the generated questions.
\textbf{(2) Our proposed framework SGSH can motivate GPT-3.5 to produce high-quality questions, which demonstrates the effectiveness of introducing skeleton heuristics.}
We observe that our approaches (i.e., SGSH(ChatGPT) and SGSH(Davinci003)) significantly outperform ChatGPT and Davinci003, because our method incorporates a novel part, i.e., skeleton heuristics, into the vanilla prompt to form a skeleton heuristics-enhanced prompt.
This prompt provides more fine-grained guidance for GPT-3.5, which can effectively guide GPT-3.5 to generate questions that are closely related to the ground-truth question. 
Furthermore, our approach (i.e., SGSH(Davinci003)) surpasses the existing baselines (i.e., Non-PLMs models and PLMs-based models), which indicates the strong capabilities of GPT-3.5 on KBQG.
\textbf{(3) Injecting skeletons into PLMs can also enhance the performance of KBQG.} 
B+S derives 8.05\% BLEU-1 gain and 10.9\% ROUGE-L gain over its corresponding vanilla model BART on WQ and obtains 6.98\% BLEU-1 gain and 4.91\% ROUGE-L gain on PQ.
This indicates the skeleton combined with the raw input can play a very significant role in guiding the question generation.

\subsection{Ablation Studies}
\label{sec:ablation_study}

\vpara{Example Selection Strategy.}
To investigate the effectiveness of our proposed skeleton-aware example selection strategy, namely ``\textit{input+skeleton emb}'', we compare it with other example selection strategies, namely ``\textit{random}'' and ``\textit{input emb}''.
Specifically, the \textit{random} signifies the random selection of examples; 
the \textit{input emb} denotes the selection of examples based on cosine similarity using the input embedding, which is derived from the last hidden state of the BART encoder;
the \textit{input+skeleton emb} introduces our innovative selection strategy that identifies examples based on the combined similarity of both the input and its corresponding skeleton.
Table~\ref{tb:ablation_study} shows the evaluation results.
Compared with other strategies, our proposed strategy (i.e., \textit{input+skeleton emb}) achieves the best performance as our strategy takes into account the proximity of the input as well as the consistency of the skeleton in the latent space, which significantly contributes to retrieving examples that are similar to the test input.
For instance, \textit{input+skeleton emb} achieves 10.15\% gain over \textit{random} and 1.09\% gain over \textit{input emb} regarding BLEU-4.

\vpara{Number of In-Context Examples.} 
To explore the effect of different numbers of in-context examples ($k$), we set $k \in \{8, 16\}$ for each test input.
As shown in Table~\ref{tb:ablation_study}, the performance of SGSH(Davinci003) improves with the increase of $k$.
For example, SGSH(Davinci003) with $k$ = 16 demonstrates better performance compared to $k$ = 8  in terms of BLEU-4 (\textbf{74.13\%} vs. 72.96\%) and ROUGE-L (\textbf{92.47\%} vs. 91.99\%).
\hide{One possible explanation is that more examples can provide more effective guidance for each test input to approach the ground-truth question.}
This suggests that providing GPT-3.5 with few-shot in-context examples is key for enabling its capability on KBQG.

\vpara{Number of Learnable Prompt Groups.}
We study the effect of different numbers of learnable prompt groups ($t$) and set $t \in \{1, 8\}$.
As $t$ increases, the performance can be significantly boosted in Table~\ref{tb:ablation_study}.
For instance, when $t$ is set to 8, the performance in BLEU-4 obtains 2.83\% gain and the performance in ROUGE-L gets 1.63\% gain compared to $t$ = 1.
It is worth noting that various groups of learnable prompts possess distinct hyperparameters, which are ensembled during the inference stage. 
This can be explained by the fact that various prompts focus on distinct aspects, thus integrating them together facilitates the KBQG.

\vpara{Proportion of Training Data for Skeleton Generator.}
As indicated in Table~\ref{tb:ablation_study} and Table~\ref{tb:overall_evaluation}, we find an interesting phenomenon that our SGSH can significantly outperform the existing SOTA model DSM using only 10\% of the training data to optimize the skeleton generator (\textbf{72.72\%} vs. 61.03\% in BLEU-4 and \textbf{91.66\%} vs. 86.06\% in ROUGE-L).
This shows the effectiveness of the skeleton generator we developed for steering GPT-3.5 toward the target question for the KBQG task with only a small amount of training data.

\begin{table}[!t]
\centering
\newcolumntype{?}{!{\vrule width 1pt}}
\renewcommand\arraystretch{1.0}
\scalebox{0.78}{
\begin{tabular}{@{}c@{ }?@{ }ccc@{ }}
\toprule
&  \textbf{BLEU-1} & \textbf{BLEU-4} & \textbf{ROUGE-L} \\
\midrule
\multicolumn{4}{c}{\textbf{Example selection strategy}}\\
\midrule
Random & 83.50 & 63.98 & 87.81\\
Input emb & 88.33 & 73.04 & 91.89\\
Input+skeleton emb & \textbf{88.87} & \textbf{74.13}& \textbf{92.47} \\
\midrule
\multicolumn{4}{c}{\textbf{Number of examples (k)}}\\
\midrule
8 & 88.16 & 72.96 & 91.99\\
16 & \textbf{88.87} & \textbf{74.13} & \textbf{92.47}\\
\midrule
\multicolumn{4}{c}{\textbf{Number of learnable prompt groups (t)}}\\
\midrule
1 & 87.36 & 71.30 & 90.84\\
8 & \textbf{88.87} &\textbf{74.13} & \textbf{92.47}\\
\midrule
\multicolumn{4}{c}{\textbf{Proportion of training data for skeleton generator }}\\
\midrule
10\% & 88.13 & 72.72& 91.66\\
100\% & \textbf{88.87} & \textbf{74.13} & \textbf{92.47}\\
\bottomrule
\end{tabular}
}
\caption{SGSH(Davinci003) ablation studies on PQ.}
\label{tb:ablation_study}
\end{table}

\begin{table}[!t]
\centering
\newcolumntype{?}{!{\vrule width 1pt}}
\renewcommand\arraystretch{1.0}
\scalebox{0.78}{
\begin{tabular}{@{}l@{ }?cc?cc@{}}
\toprule
\multirow{2}{*} {\textbf{Model}} &
\multicolumn{2}{c?}{\textbf{GRAFT-Net}} &
\multicolumn{2}{c}{\textbf{NSM}}
\\
& \textbf{F1}& \textbf{Hits@1}& \textbf{F1}& \textbf{Hits@1}\\
\midrule
Real & \textbf{0.622} & \textbf{0.681} & \textbf{0.666} & \textbf{0.727} \\
\midrule
-o & 0.493 & 0.575 & 0.524 & 0.594\\
+DSM & 0.604 &0.664& 0.663 & 0.721 \\
+B+S &0.606 & 0.676 &0.664& 0.724\\
+SGSH(Davinci003) & 0.618 & 0.677 &\textbf{0.666} & 0.726 \\
\bottomrule
\end{tabular}
}
\caption{QA performance of GRAFT-Net and NSM.}
\vspace{-0.3cm}
\label{tab:QA}
\end{table}

\subsection{Effect on QA Performance }
\label{sec:qa}
We explore whether our SGSH can contribute to QA tasks.
GRAFT-Net~\cite{GRAFT-Net} and NSM~\cite{He2021} are two popular KBQA models utilized for experiments on WebQSP~\cite{WQ}, a widely adopted KBQA dataset with 2,848 (question, answer) training instances.
There are 1,409 (question, answer) pairs in WebQSP overlapping with WQ. Then we can quickly obtain their corresponding subgraphs from WQ, so we conduct experiments by replacing some of the (question, answer) pairs in WebQSP with questions generated by KBQG models on WQ.
Specifically, we train GRAFT-Net and NSM on the datasets partially replaced by the pseudo questions generated by DSM, B+S, and SGSH(Davinci003), denoting them as ``+DSM'', ``+B+S'', and ``+SGSH(Davinci003)'' respectively.
We also train GRAFT-Net and NSM on the original WebQSP, denoted as ``Real'', and a modified version where overlapping instances are eliminated, indicated as ``-o''.
Finally, we compare their performances with Real.

Table~\ref{tab:QA} reports F1 and Hits@1 of GRAFT-Net and NSM on various datasets.
From the results, we can draw the following conclusions.
\textbf{(1) The generated questions and the corresponding answers form (question, answer) pairs which can be seen as a data augmentation method for KBQA,} 
because GRAFT-Net and NSM perform better than -o on +DSM, +B+S, and +SGSH(Davinci003).
\textbf{(2) SGSH(Davinci003) generates better questions than other baselines (i.e., DSM, B+S),} because +SGSH(Davinci003) outperforms all other baselines. 
\textbf{(3) The questions generated by SGSH(Davinci003) closely resemble the actual questions,} because +SGSH(Davinci003) and Real have comparable results.

\subsection{Human Evaluation}
\label{sec:human}
To further explore the effectiveness of SGSH, we randomly select 50 test examples $\mathcal{S}_{50} = \{(G_j,a_j, q_j)\}_{j=1}^{50}$ from WQ.
Then we assess the generated questions from three perspectives: fluency, relevance, and diversity.
Fluency aims to evaluate whether the generated questions are human-readable. Relevance measures how relevant the generated question is to the input.
Meanwhile, diversity focuses on assessing the extent to which the generated questions differ from the ground truth.
We use the five-point Likert scale to score fluency, relevance, and diversity, where 1 is a poor score and 5 is a perfect score.
We invite three persons to score all questions generated by our SGSH(Davinci003) and two baselines (i.e., DSM and B+S), and then average their scores as the final result.
As shown in Table~\ref{tb:human_evaluation}, our proposed SGSH consistently outperforms other baselines in fluency, relevance, and diversity.
Besides, our method is comparable to the ground-truth question in fluency and relevance.

\section{Related Work}

\vpara{Knowledge Base Question Generation (KBQG).}
KBQG has evolved significantly over recent years, primarily driven by advancements in sequence-to-sequence (Seq2Seq) modeling approaches \cite{ bi2020knowledge, chen2020toward, kumar2019difficulty, liu2019generating}. Early models focused on encoding serialized subgraphs and specific answers into intermediate representations, which were then decoded into questions. These initial methods, while effective, were limited by the scope of their training data. This limitation paved the way for pre-trained language models like BART~\cite{BART} and T5~\cite{T5}, which brought a paradigm shift in KBQG \cite{Diversify_Coling24, DSM, JointGT}.
Additionally, LLMs exhibit considerable potential as they possess a substantial quantity of parameters and demonstrate impressive performance on a wide range of downstream tasks such as KBQA~\cite{LLM_KBQA} and fact-checking~\cite{fact-check}.
However, despite these advancements, a gap remained in harnessing the full potential of LLMs like InstructGPT \cite{InstructGPT} and ChatGPT for KBQG tasks. These LLMs, with their extensive parameterization, encode a wealth of generalized knowledge but have been underutilized in the specific domain of KBQG. Concurrently to our work, KQG-COT~\cite{KQG_COT} uses unlabeled data to craft prompts to generate questions.

\vpara{In-Context Learning.}
The emergence of LLMs introduced a novel capability—In-Context Learning (ICL). ICL enables models like GPT-3.5 to adapt to new tasks through carefully designed prompts, incorporating task descriptions and relevant examples, without necessitating further parameter tuning~\cite{ICL_NIPS}. Research in ICL has unraveled intriguing insights, such as its dependency on example selection strategies and prompt templates \citet{ICL_ZHAO}, its insensitivity to ground-truth labels \citet{EMNLP22_icl}, and its unique modalities of Task Recognition and Task Learning \citet{ACL23_Chen}. Yet, the application of ICL in KBQG, especially in the context of utilizing large language models for nuanced and accurate question generation from knowledge bases, remains an underexplored area.
\section{Conclusion}

\begin{table}[!t]
\centering
\newcolumntype{?}{!{\vrule width 1pt}}
\renewcommand\arraystretch{1.0}
\scalebox{0.75}{
\begin{tabular}{@{}c@{ }?c?c?c@{}}
\toprule
\textbf{Model}  & \textbf{Fluency} &  \textbf{Relevance} &  \textbf{Diversity}\\
\midrule
DSM & 4.17 & 4.16 & 3.62\\
B+S & 4.21 & 4.18 & 3.56\\
SGSH(Davinci003) & 4.25 & 4.21 & 3.81\\
\midrule
Ground-truth & \textbf{4.39} & \textbf{4.25} &  - \\
\bottomrule
\end{tabular}
}
\caption{Human evaluation results on WQ.}
\vspace{-0.3cm}
\label{tb:human_evaluation}
\end{table}

We explore how to steer GPT-3.5 toward the gold question on KBQG.
In this paper, we propose a simple but effective framework SGSH to \underline{\textbf{S}}timulate \underline{\textbf{G}}PT-3.5 with \underline{\textbf{S}}keleton \underline{\textbf{H}}euristics for KBQG, which provides fine-grained guidance for GPT-3.5 to generate high-quality questions.
Specifically, we employ a BART-based skeleton generator that is trained on our constructed training dataset using a learnable prompting strategy to obtain skeleton heuristics.
Toward these skeleton heuristics, we then devise a skeleton heuristics-enhanced prompt to trigger GPT-3.5 to align with the ground-truth question.
Extensive experiments on widely used datasets demonstrate the advanced performance of our proposed SGSH.
In addition, optimizing the skeleton generator with only a small amount of training data (i.e., 10\%) can outperform existing SOTA (i.e., DSM).
This fine-grained guiding framework could be inspiring for other NLP tasks.

\section*{Acknowledgments}
This work is supported by National Key Research \& Develop Plan (2023YFF0725100) and the National Natural Science Foundation of China (62322214, U23A20299, 62076245, 62072460, 62172424, 62276270). This work is supported by Public Computing Cloud, Renmin University of China.  We also acknowledge the support from the China Scholarship Council Scholarship Fund. We are sincerely grateful to all reviewers for their insightful feedback.

\section*{Limitations}
In this section, we discuss the limitations of this work from two aspects.
Firstly, the effectiveness of our method is influenced by the accuracy of the skeleton heuristics. The scarcity of labeled data for training the skeleton generator has motivated us to explore automatic training data construction, utilizing both rule-based and ChatGPT-based approaches. However, the quality of this synthetically produced training data is inherently constrained by the capabilities of ChatGPT.
Secondly, a more diverse range of datasets for evaluating the generalizability of KBQG models is under-explored. 
In future work, we plan to establish a comprehensive benchmark dataset encompassing a broad spectrum of domains. This benchmark will enable a more detailed evaluation of our approach and contribute significantly to the ongoing development within the KBQG community.

\bibliography{aaai24}

\begin{thebibliography}{36}
\expandafter\ifx\csname natexlab\endcsname\relax\def\natexlab#1{#1}\fi

\bibitem[{Baek et~al.(2023)Baek, Aji, and Saffari}]{LLM_KBQA}
Jinheon Baek, Alham~Fikri Aji, and Amir Saffari. 2023.
\newblock Knowledge-augmented language model prompting for zero-shot knowledge graph question answering.
\newblock \emph{arXiv preprint arXiv:2306.04136}.

\bibitem[{Bi et~al.(2020)Bi, Cheng, Li, Wang, and Qi}]{bi2020knowledge}
Sheng Bi, Xiya Cheng, Yuan{-}Fang Li, Yongzhen Wang, and Guilin Qi. 2020.
\newblock Knowledge-enriched, type-constrained and grammar-guided question generation over knowledge bases.
\newblock In \emph{Proceedings of the 28th International Conference on Computational Linguistics, {COLING} 2020}, pages 2776--2786.

\bibitem[{Brown et~al.(2020)Brown, Mann, Ryder, Subbiah, Kaplan, Dhariwal, Neelakantan, Shyam, Sastry, Askell, Agarwal, Herbert{-}Voss, Krueger, Henighan, Child, Ramesh, Ziegler, Wu, Winter, Hesse, Chen, Sigler, Litwin, Gray, Chess, Clark, Berner, McCandlish, Radford, Sutskever, and Amodei}]{ICL_NIPS}
Tom~B. Brown, Benjamin Mann, Nick Ryder, Melanie Subbiah, Jared Kaplan, Prafulla Dhariwal, Arvind Neelakantan, Pranav Shyam, Girish Sastry, Amanda Askell, Sandhini Agarwal, Ariel Herbert{-}Voss, Gretchen Krueger, Tom Henighan, Rewon Child, Aditya Ramesh, Daniel~M. Ziegler, Jeffrey Wu, Clemens Winter, Christopher Hesse, Mark Chen, Eric Sigler, Mateusz Litwin, Scott Gray, Benjamin Chess, Jack Clark, Christopher Berner, Sam McCandlish, Alec Radford, Ilya Sutskever, and Dario Amodei. 2020.
\newblock Language models are few-shot learners.
\newblock In \emph{Proceedings of the 34th Conference on Neural Information Processing Systems, {NeurIPS} 2020}, pages 1877--1901.

\bibitem[{Chen et~al.(2023)Chen, Wu, and Zaki}]{chen2020toward}
Yu~Chen, Lingfei Wu, and Mohammed~J. Zaki. 2023.
\newblock Toward subgraph-guided knowledge graph question generation with graph neural networks.
\newblock \emph{{IEEE} Transactions on Neural Networks and Learning Systems}, pages 1--12.

\bibitem[{Finn et~al.(2017)Finn, Abbeel, and Levine}]{MAML}
Chelsea Finn, Pieter Abbeel, and Sergey Levine. 2017.
\newblock Model-agnostic meta-learning for fast adaptation of deep networks.
\newblock In \emph{Proceedings of the 34th International Conference on Machine Learning, {ICML} 2017}, pages 1126--1135.

\bibitem[{Guo et~al.(2024)Guo, Liao, Li, and Chua}]{guo2024survey}
Shasha Guo, Lizi Liao, Cuiping Li, and Tat-Seng Chua. 2024.
\newblock A survey on neural question generation: Methods, applications, and prospects.
\newblock \emph{arXiv preprint arXiv:2402.18267}.

\bibitem[{Guo et~al.(2023)Guo, Zhang, Ke, Li, and Chen}]{Diversify_Coling24}
Shasha Guo, Jing Zhang, Xirui Ke, Cuiping Li, and Hong Chen. 2023.
\newblock Diversifying question generation over knowledge base via external natural questions.
\newblock \emph{arXiv preprint arXiv:2309.14362}.

\bibitem[{Guo et~al.(2022)Guo, Zhang, Wang, Zhang, Li, and Chen}]{DSM}
Shasha Guo, Jing Zhang, Yanling Wang, Qianyi Zhang, Cuiping Li, and Hong Chen. 2022.
\newblock Dsm: Question generation over knowledge base via modeling diverse subgraphs with meta-learner.
\newblock In \emph{Proceedings of the 2022 Conference on Empirical Methods in Natural Language Processing, {EMNLP} 2022}, pages 4194--4207.

\bibitem[{He et~al.(2021)He, Lan, Jiang, Zhao, and Wen}]{He2021}
Gaole He, Yunshi Lan, Jing Jiang, Wayne~Xin Zhao, and Ji{-}Rong Wen. 2021.
\newblock Improving multi-hop knowledge base question answering by learning intermediate supervision signals.
\newblock In \emph{Proceedings of the 14th {ACM} International Conference on Web Search and Data Mining, {WSDM} 21}, pages 553--561.

\bibitem[{Ke et~al.(2021)Ke, Ji, Ran, Cui, Wang, Song, Zhu, and Huang}]{JointGT}
Pei Ke, Haozhe Ji, Yu~Ran, Xin Cui, Liwei Wang, Linfeng Song, Xiaoyan Zhu, and Minlie Huang. 2021.
\newblock Jointgt: Graph-text joint representation learning for text generation from knowledge graphs.
\newblock In \emph{Findings of the Association for Computational Linguistics: {ACL/IJCNLP} 2021}, pages 2526--2538.

\bibitem[{Kumar et~al.(2019)Kumar, Hua, Ramakrishnan, Qi, Gao, and Li}]{kumar2019difficulty}
Vishwajeet Kumar, Yuncheng Hua, Ganesh Ramakrishnan, Guilin Qi, Lianli Gao, and Yuan-Fang Li. 2019.
\newblock Difficulty-controllable multi-hop question generation from knowledge graphs.
\newblock In \emph{Proceddings of the 18th International Semantic Web Conference, {ISWC} 2019}, pages 382--398.

\bibitem[{Lester et~al.(2021)Lester, Al{-}Rfou, and Constant}]{prompt_tuning}
Brian Lester, Rami Al{-}Rfou, and Noah Constant. 2021.
\newblock The power of scale for parameter-efficient prompt tuning.
\newblock In \emph{Proceedings of the 2021 Conference on Empirical Methods in Natural Language Processing, {EMNLP} 2021}, pages 3045--3059.

\bibitem[{Lewis et~al.(2020)Lewis, Liu, Goyal, Ghazvininejad, Mohamed, Levy, Stoyanov, and Zettlemoyer}]{BART}
Mike Lewis, Yinhan Liu, Naman Goyal, Marjan Ghazvininejad, Abdelrahman Mohamed, Omer Levy, Veselin Stoyanov, and Luke Zettlemoyer. 2020.
\newblock {BART:} denoising sequence-to-sequence pre-training for natural language generation, translation, and comprehension.
\newblock In \emph{Proceedings of the 58th Annual Meeting of the Association for Computational Linguistics, {ACL} 2020}, pages 7871--7880.

\bibitem[{Li et~al.(2023)Li, Peng, and Zhang}]{fact-check}
Miaoran Li, Baolin Peng, and Zhu Zhang. 2023.
\newblock Self-checker: Plug-and-play modules for fact-checking with large language models.
\newblock \emph{arXiv preprint arXiv:2305.14623}.

\bibitem[{Liang and Liao(2023)}]{ClusterPrompt}
Jinggui Liang and Lizi Liao. 2023.
\newblock Clusterprompt: Cluster semantic enhanced prompt learning for new intent discovery.
\newblock In \emph{Findings of the Association for Computational Linguistics: {EMNLP} 2023, Singapore, December 6-10, 2023}, pages 10468--10481.

\bibitem[{Liang et~al.(2023)Liang, Wang, Zhu, Wang, Qian, and Lan}]{KQG_COT}
Yuanyuan Liang, Jianing Wang, Hanlun Zhu, Lei Wang, Weining Qian, and Yunshi Lan. 2023.
\newblock Prompting large language models with chain-of-thought for few-shot knowledge base question generation.
\newblock In \emph{Proceedings of the 2023 Conference on Empirical Methods in Natural Language Processing, {EMNLP} 2023}, pages 4329--4343.

\bibitem[{Lin and Och(2004)}]{rouge}
Chin-Yew Lin and FJ~Och. 2004.
\newblock Looking for a few good metrics: Rouge and its evaluation.
\newblock In \emph{Ntcir workshop}, pages 1--8.

\bibitem[{Liu et~al.(2019)Liu, Liu, He, Nie, and Zhao}]{liu2019generating}
Cao Liu, Kang Liu, Shizhu He, Zaiqing Nie, and Jun Zhao. 2019.
\newblock Generating questions for knowledge bases via incorporating diversified contexts and answer-aware loss.
\newblock In \emph{Proceedings of the 2019 Conference on Empirical Methods in Natural Language Processing and the 9th International Joint Conference on Natural Language Processing, {EMNLP-IJCNLP} 2019}, pages 2431--2441.

\bibitem[{Liu et~al.(2023)Liu, Bao, Zhang, Zhang, Hu, Zhang, and Yan}]{codeGeneration}
Chao Liu, Xuanlin Bao, Hongyu Zhang, Neng Zhang, Haibo Hu, Xiaohong Zhang, and Meng Yan. 2023.
\newblock Improving chatgpt prompt for code generation.
\newblock \emph{arXiv preprint arXiv:2305.08360}.

\bibitem[{Liu et~al.(2022)Liu, Shen, Zhang, Dolan, Carin, and Chen}]{in_context_gpt}
Jiachang Liu, Dinghan Shen, Yizhe Zhang, Bill Dolan, Lawrence Carin, and Weizhu Chen. 2022.
\newblock What makes good in-context examples for gpt-3?
\newblock In \emph{Proceedings of Deep Learning Inside Out: The 3rd Workshop on Knowledge Extraction and Integration for Deep Learning Architectures, {DeeLIO@ACL} 2022}, pages 100--114.

\bibitem[{Min et~al.(2022{\natexlab{a}})Min, Lyu, Holtzman, Artetxe, Lewis, Hajishirzi, and Zettlemoyer}]{in_context}
Sewon Min, Xinxi Lyu, Ari Holtzman, Mikel Artetxe, Mike Lewis, Hannaneh Hajishirzi, and Luke Zettlemoyer. 2022{\natexlab{a}}.
\newblock Rethinking the role of demonstrations: What makes in-context learning work?
\newblock In \emph{Proceedings of the 2022 Conference on Empirical Methods in Natural Language Processing, {EMNLP} 2022}, pages 11048--11064.

\bibitem[{Min et~al.(2022{\natexlab{b}})Min, Lyu, Holtzman, Artetxe, Lewis, Hajishirzi, and Zettlemoyer}]{EMNLP22_icl}
Sewon Min, Xinxi Lyu, Ari Holtzman, Mikel Artetxe, Mike Lewis, Hannaneh Hajishirzi, and Luke Zettlemoyer. 2022{\natexlab{b}}.
\newblock Rethinking the role of demonstrations: What makes in-context learning work?
\newblock In \emph{Proceedings of the 2022 Conference on Empirical Methods in Natural Language Processing, {EMNLP} 2022}, pages 11048--11064.

\bibitem[{Nan et~al.(2023)Nan, Zhao, Zou, Ri, Tae, Zhang, Cohan, and Radev}]{Text-to-SQL}
Linyong Nan, Yilun Zhao, Weijin Zou, Narutatsu Ri, Jaesung Tae, Ellen Zhang, Arman Cohan, and Dragomir Radev. 2023.
\newblock Enhancing few-shot text-to-sql capabilities of large language models: A study on prompt design strategies.
\newblock \emph{arXiv preprint arXiv:2305.12586}.

\bibitem[{Ouyang et~al.(2022)Ouyang, Wu, Jiang, Almeida, Wainwright, Mishkin, Zhang, Agarwal, Slama, Ray, Schulman, Hilton, Kelton, Miller, Simens, Askell, Welinder, Christiano, Leike, and Lowe}]{InstructGPT}
Long Ouyang, Jeffrey Wu, Xu~Jiang, Diogo Almeida, Carroll~L. Wainwright, Pamela Mishkin, Chong Zhang, Sandhini Agarwal, Katarina Slama, Alex Ray, John Schulman, Jacob Hilton, Fraser Kelton, Luke Miller, Maddie Simens, Amanda Askell, Peter Welinder, Paul~F. Christiano, Jan Leike, and Ryan Lowe. 2022.
\newblock Training language models to follow instructions with human feedback.
\newblock In \emph{Proceedings of the 36th Conference on Neural Information Processing Systems, {NeurIPS} 2022}, pages 27730--27744.

\bibitem[{Pan et~al.(2023)Pan, Gao, Chen, and Chen}]{ACL23_Chen}
Jane Pan, Tianyu Gao, Howard Chen, and Danqi Chen. 2023.
\newblock What in-context learning "learns" in-context: Disentangling task recognition and task learning.
\newblock In \emph{Findings of the Association for Computational Linguistics: {ACL} 2023}, pages 8298--8319.

\bibitem[{Papineni et~al.(2002)Papineni, Roukos, Ward, and Zhu}]{bleu}
Kishore Papineni, Salim Roukos, Todd Ward, and Wei-Jing Zhu. 2002.
\newblock Bleu: a method for automatic evaluation of machine translation.
\newblock In \emph{Proceedings of the 40th annual meeting of the Association for Computational Linguistics, {ACL} 2002}, pages 311--318.

\bibitem[{Raffel et~al.(2020)Raffel, Shazeer, Roberts, Lee, Narang, Matena, Zhou, Li, and Liu}]{T5}
Colin Raffel, Noam Shazeer, Adam Roberts, Katherine Lee, Sharan Narang, Michael Matena, Yanqi Zhou, Wei Li, and Peter~J. Liu. 2020.
\newblock Exploring the limits of transfer learning with a unified text-to-text transformer.
\newblock \emph{J. Mach. Learn. Res.}, 21:140:1--140:67.

\bibitem[{Saeidi et~al.(2018)Saeidi, Bartolo, Lewis, Singh, Rockt{\"{a}}schel, Sheldon, Bouchard, and Riedel}]{ConvMR}
Marzieh Saeidi, Max Bartolo, Patrick S.~H. Lewis, Sameer Singh, Tim Rockt{\"{a}}schel, Mike Sheldon, Guillaume Bouchard, and Sebastian Riedel. 2018.
\newblock Interpretation of natural language rules in conversational machine reading.
\newblock In \emph{Proceedings of the 2018 Conference on Empirical Methods in Natural Language Processing, {EMNLP} 2018}, pages 2087--2097.

\bibitem[{Sun et~al.(2018)Sun, Dhingra, Zaheer, Mazaitis, Salakhutdinov, and Cohen}]{GRAFT-Net}
Haitian Sun, Bhuwan Dhingra, Manzil Zaheer, Kathryn Mazaitis, Ruslan Salakhutdinov, and William~W. Cohen. 2018.
\newblock Open domain question answering using early fusion of knowledge bases and text.
\newblock In \emph{Proceedings of the 2018 Conference on Empirical Methods in Natural Language Processing, {EMNLP} 2018}, pages 4231--4242.

\bibitem[{Talmor and Berant(2018)}]{CWQ}
Alon Talmor and Jonathan Berant. 2018.
\newblock The web as a knowledge-base for answering complex questions.
\newblock In \emph{Proceedings of the 2018 Conference of the North American Chapter of the Association for Computational Linguistics: Human Language Technologies, {NAACL-HLT} 2018}, pages 641--651.

\bibitem[{Vaswani et~al.(2017)Vaswani, Shazeer, Parmar, Uszkoreit, Jones, Gomez, Kaiser, and Polosukhin}]{vaswani2017attention}
Ashish Vaswani, Noam Shazeer, Niki Parmar, Jakob Uszkoreit, Llion Jones, Aidan~N Gomez, {\L}ukasz Kaiser, and Illia Polosukhin. 2017.
\newblock Attention is all you need.
\newblock In \emph{Proceedings of the 31st Conference on Neural Information Processing Systems, {NeurIPS} 2017}, pages 5998--6008.

\bibitem[{Wang et~al.(2021)Wang, Lan, and Baraniuk}]{MWPG}
Zichao Wang, Andrew~S. Lan, and Richard~G. Baraniuk. 2021.
\newblock Math word problem generation with mathematical consistency and problem context constraints.
\newblock In \emph{Proceedings of the 2021 Conference on Empirical Methods in Natural Language Processing, {EMNLP} 2021}, pages 5986--5999.

\bibitem[{Xiong et~al.(2022)Xiong, Bao, Zhao, Wu, and He}]{AutoQGS}
Guanming Xiong, Junwei Bao, Wen Zhao, Youzheng Wu, and Xiaodong He. 2022.
\newblock Autoqgs: Auto-prompt for low-resource knowledge-based question generation from {SPARQL}.
\newblock In \emph{Proceedings of the 31st {ACM} International Conference on Information {\&} Knowledge Management}, pages 2250--2259.

\bibitem[{Yih et~al.(2016)Yih, Richardson, Meek, Chang, and Suh}]{WQ}
Wen-tau Yih, Matthew Richardson, Christopher Meek, Ming-Wei Chang, and Jina Suh. 2016.
\newblock The value of semantic parse labeling for knowledge base question answering.
\newblock In \emph{Proceedings of the 54th Annual Meeting of the Association for Computational Linguistics, {ACL} 2016}, pages 201--206.

\bibitem[{Zhao et~al.(2021)Zhao, Wallace, Feng, Klein, and Singh}]{ICL_ZHAO}
Zihao Zhao, Eric Wallace, Shi Feng, Dan Klein, and Sameer Singh. 2021.
\newblock Calibrate before use: Improving few-shot performance of language models.
\newblock In \emph{Proceedings of the 38th International Conference on Machine Learning, {ICML} 2021}, pages 12697--12706.

\bibitem[{Zhou et~al.(2018)Zhou, Huang, and Zhu}]{PQ}
Mantong Zhou, Minlie Huang, and Xiaoyan Zhu. 2018.
\newblock An interpretable reasoning network for multi-relation question answering.
\newblock In \emph{Proceedings of the 27th International Conference on Computational Linguistics, {COLING} 2018}, pages 2010--2022.

\end{thebibliography}
\appendix
\begin{algorithm}[!t]	\renewcommand{\algorithmicrequire}
    {\textbf{Input:}}
	\renewcommand{\algorithmicensure}{\textbf{Output:}}
	\renewcommand{\algorithmicreturn}{\textbf{Return}}
	\caption{\textbf{Skeleton generator training}}
	\label{alg:skeleton_generator}
	\begin{algorithmic}[1]
	    \REQUIRE $\mathcal{D}=\{ (G_{i}, a_{i}, q_i)\}_{i=1}^N$, Training epochs $T$.
	    \ENSURE Parameters of skeleton generator $f_{\text{PLM}}$.
        \STATE Initialize the skeleton set $S$ = $\varnothing$;
        \FOR{each $q_i \in \mathcal{D}$}
	    \STATE Extract the skeleton $s_i^{\prime}$ using rule-based skeleton extractor;
	    \STATE Generate the skeleton $s_i^{\prime \prime}$ using ChatGPT-based skeleton generator;
            \STATE Score $s_i^{\prime}$ and $s_i^{\prime \prime}$ using ChatGPT-based skeleton quality evaluator;
            \STATE Obtain refined skeleton $s_i$ = MaxScore ($s_i^{\prime}$, $s_i^{\prime \prime}$);
            \STATE $S$ = $S \cup \{s_i\}$;
	\ENDFOR
       \STATE Acquire supervised data $\mathcal{D_S} =  \{(G_i, a_i, s_i)\}_{i=1}^N$ based on $\mathcal{D}$ and $S$ to train $f_{\text{PLM}}$;
       \STATE Initialize parameters of learnable prompts $\theta_{p}$ and parameters of backbone BART $\theta$;
       \FOR{$epoch \leftarrow 1$ \textbf{to} $T$}
        \STATE Calculate $\mathcal{L}(\theta, \theta_{p})$ via Eq.1;
    
        \STATE $\theta, \theta_{p} \leftarrow$ AdamW($\theta, \theta_{p}, \mathcal{L}$);
       \ENDFOR
        \STATE Return $\theta$ and $\theta_{p}$      
	\end{algorithmic}  
\end{algorithm}

\section{Appendix}
\label{sec:app}

\subsection{Skeleton Generator}

\vpara{Training Process.}
For our proposed SGSH, the skeleton generator $f_{\text{PLM}}$ is a crucial module to obtain skeleton heuristics.
Algorithm~\ref{alg:skeleton_generator} details the training process of our devised skeleton generator.
We present the automatic training data construction strategy to construct labeled data $\mathcal{D_S} =  \{(G_i, a_i, s_i)\}_{i=1}^N$ for training skeleton generator $f_{\text{PLM}}$ in Lines 1-9.
More specifically, we utilize a rule-based approach to extract the skeleton $s_{i}^{\prime}$ from the target question $q_i$ on $\mathcal{D}$ by searching a pre-defined vocabulary of skeleton elements (Line 3).
We employ the powerful potential of ChatGPT to generate the skeleton $s_{i}^{\prime \prime}$ for the target question $q_i$ on $\mathcal{D}$ (Line 4).
Subsequently, we use ChatGPT as an automatic grader to score $s_{i}^{\prime}$ and $s_{i}^{\prime \prime}$ (Lines 5-6). 
We choose the higher score one as the refined skeleton $s_i$ to put into the skeleton set $S$ (Line 7).
Based on obtained labeled data $\mathcal{D_S} =  \{(G_i, a_i, s_i)\}_{i=1}^N$, 
we apply a learnable prompting strategy to train skeleton generator $f_{\text{PLM}}$ (Lines 10-13).

\begin{figure}[!t]
\centering 
\includegraphics[width=0.47\textwidth]{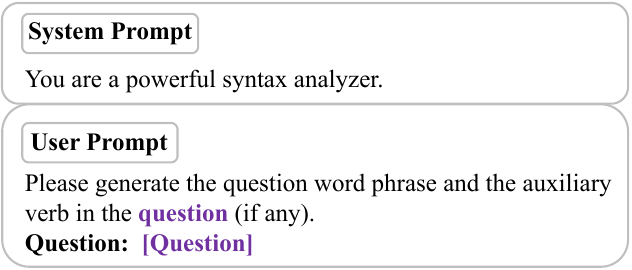}
\caption{A ChatGPT prompt for generating skeletons.}
\label{fig:prompt_chat_generator} 
\end{figure}

\begin{figure}[!t]
\centering 
\includegraphics[width=0.47\textwidth]{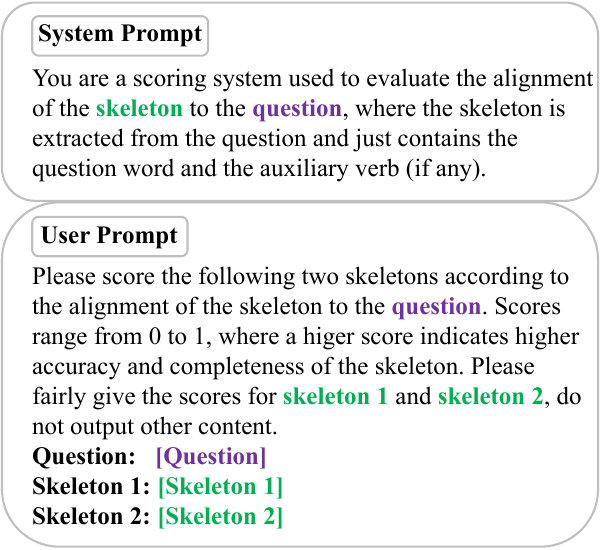}
\caption{A ChatGPT prompt for scoring and selecting high-quality skeletons.}
\label{fig:prompt_chat_scorer} 
\end{figure}

\vpara{Prompt of ChatGPT-based Skeleton Generator.}\
The rule-based method retrieves the skeleton (i.e., the question word phrase and the auxiliary verb) from the target question through a search within a pre-defined vocabulary of skeleton elements.
Obviously, its challenge is in addressing complex questions with nested clauses.
Hence, we leverage the capabilities of ChatGPT as an enhanced skeleton generator to generate skeletons for target questions, especially those that are inherently complex.
Figure~\ref{fig:prompt_chat_generator} demonstrates the system and user prompts of the ChatGPT-based skeleton generator.

\vpara{Prompt of ChatGPT-based Skeleton Quality Evaluator.}
To circumvent the costly and time-consuming human selection, we exploit the potential of ChatGPT as an automatic scorer. 
This capability empowers us to effectively filter out low-scoring skeletons generated by the rule-based skeleton extractor and the ChatGPT-based skeleton generator.
Figure~\ref{fig:prompt_chat_scorer} illustrates the system and user prompts of the ChatGPT-based skeleton quality evaluator. ``Question'' denotes the specific target question, ``Skeleton 1'' corresponds to the skeleton extracted through the rule-based approach, and ``Skeleton 2'' represents the skeleton generated by the ChatGPT-based generator.

\begin{figure}[t!]
\centering 
\includegraphics[width=0.47\textwidth]{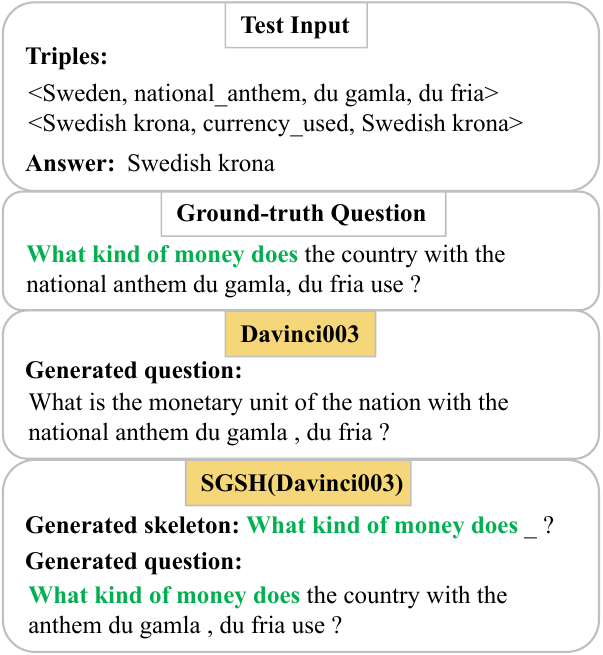}
\caption{Illustration of an example from WQ dataset, which shows the question generated by Davinci003 and our method SGSH(Davinci003).}
\label{fig:wq_example} 
\end{figure}

\begin{figure}[t!]
\centering 
\includegraphics[width=0.47\textwidth]{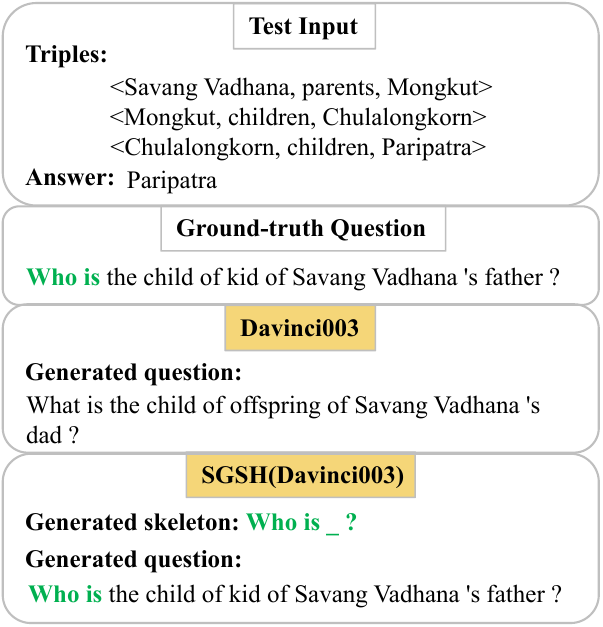}
\caption{Illustration of an example from PQ dataset, which presents the question generated by Davinci003 and our approach SGSH(Davinci003).}
\label{fig:pq_example} 
\end{figure}

\subsection{Experimental Implementation Details}
\label{sec:append_exp}

\vpara{Code Implementation.} We implement our method in Pytorch, 
and run all experiments on a server with a single Nvidia RTX A6000 (48G) GPU card, an Intel(R) Xeon(R) Gold 5218R CPU, 256GB memory, and the Ubuntu 20.04.2 LTS operating system.

\vpara{Skeleton Generator $f_{\text{PLM}}$.} We employ BART-base\footnote{https://huggingface.co/facebook/bart-base} as the backbone of the skeleton generator and fine-tune it with the AdamW optimizer.
We set the learning rate as 5e-5, batch size as 16, training epochs as 20. 
We initialize the learnable prompts from word embeddings in the vocabulary. 
We compare different lengths of the prompt such as [2, 4, 8, 16, 32], and set it to 16.
We train 8 groups of learnable prompts using different learning rates (abbreviated as lr) and batch sizes (abbreviated as bs) including $f_{\text{PLM}}$ (lr = 2e-5, bs = 8), $f_{\text{PLM}}$ (lr = 2e-5, bs = 16), $f_{\text{PLM}}$ (lr = 3e-5, bs = 8), $f_{\text{PLM}}$ (lr = 3e-5, bs = 16), $f_{\text{PLM}}$ (lr = 4e-5, bs = 8), $f_{\text{PLM}}$ (lr = 4e-5, bs = 16), $f_{\text{PLM}}$ (lr = 5e-5, bs = 8), and $f_{\text{PLM}}$ (lr = 5e-5, bs = 16).

\vpara{Frozen GPT-3.5 Model.} We utilize two versions of the  GPT-3.5 series models including \texttt{text-davinci-003} (i.e., Davinci003) and \texttt{gpt-3.5-turbo} (i.e., ChatGPT) in our experiments.
We use our proposed skeleton-aware example selection strategy (i.e., ``\textit{input+skeleton emb}'') to choose in-context examples and set the number of these examples as 16.
We set n as 10 and employ a majority voting approach across the n questions to determine the final question. 
We set temperature as 0.7, top\_p as 1, frequency\_penalty as 0, and presence\_penalty as 0.

\subsection{Running Examples}
We provide two illustrative examples for the WQ and PQ datasets in Figure~\ref{fig:wq_example} and Figure~\ref{fig:pq_example}, respectively.
For each example, we present the generated questions by Davinci003 and our approach SGSH(Davinci003).
we observe that: (1) The questions generated by SGSH(Davinci003) are more closely related to the ground-truth questions compared to Davinci003, which shows the superiority of our devised framework.
(2) The skeletons produced by the skeleton generator are similar to the actual skeletons of the ground-truth question, which demonstrates the effectiveness of our devised skeleton generator.
(3) The questions generated by Davinci003 express similar semantics to ground-truth questions but differ in surface form, which verifies that Davinci003 contains rich semantic knowledge, but requires more fine-grained guidance to stimulate it toward the ground-truth question.
Nevertheless, our proposed SGSH framework effectively addresses this challenge.

\end{document}